\documentclass[letterpaper, 10 pt, conference]{ieeeconf}  

\IEEEoverridecommandlockouts                              	

\usepackage[pdftex]{graphicx}
\usepackage{float}
\usepackage{multirow}
\usepackage{amsmath}
\usepackage{amssymb}					
\usepackage{subcaption}
\usepackage{algpseudocode}
\usepackage{algorithm}
\usepackage{romannum}
\usepackage{paralist}
\usepackage[style=numeric,giveninits=true, sorting=none]{biblatex}

\title{\LARGE \bf
HDPV-SLAM: Hybrid Depth-augmented Panoramic Visual SLAM for Mobile Mapping System with Tilted LiDAR and Panoramic Visual Camera}

\author{Mostafa Ahmadi$^{1}$, Amin Alizadeh Naeini$^{1}$, Mohammad Moein Sheikholeslami$^{1}$, Zahra Arjmandi$^{1}$, \\ Yujia Zhang$^{1}$, and Gunho Sohn$^{1}$$^{\dagger}$
\thanks{$^{\dagger}$Corresponding author}
\thanks{$^{1}$The authors are 
        with the Department of Earth and Space Science and Engineering, 
        Lassonde School of Engineering, 
        York University, 4700 Keele Street, Toronto, Ontario M3J 1P3, Canada.
        {\t\small ahmadism@yorku.ca, naeini@yorku.ca, mmoein@yorku.ca, zahraarj@yorku.com, zhang89@yorku.ca, gsohn@yorku.ca}}
}

\begin{document}

\maketitle
\thispagestyle{empty}
\pagestyle{empty}

\begin{abstract}
This paper proposes a novel visual simultaneous localization and mapping (SLAM) system called Hybrid Depth-augmented Panoramic Visual SLAM (HDPV-SLAM), that employs a panoramic camera and a tilted multi-beam LiDAR scanner to generate accurate and metrically-scaled trajectories. RGB-D SLAM was the design basis for HDPV-SLAM, which added depth information to visual features. It aims to solve the two major issues hindering the performance of similar SLAM systems. The first obstacle is the sparseness of LiDAR depth, which makes it difficult to correlate it with the extracted visual features of the RGB image. A deep learning-based depth estimation module for iteratively densifying sparse LiDAR depth was suggested to address this issue. The second issue pertains to the difficulties in depth association caused by a lack of horizontal overlap between the panoramic camera and the tilted LiDAR sensor. To surmount this difficulty, we present a hybrid depth association module that optimally combines depth information estimated by two independent procedures, feature-based triangulation and depth estimation. During a phase of feature tracking, this hybrid depth association module aims to maximize the use of more accurate depth information between the triangulated depth with visual features tracked and the deep learning-based corrected depth. We evaluated the efficacy of HDPV-SLAM using the 18.95 km-long York University and Teledyne Optech (YUTO) MMS dataset. The experimental results demonstrate that the two proposed modules contribute substantially to the performance of HDPV-SLAM, which surpasses that of the state-of-the-art (SOTA) SLAM systems.
\end{abstract}

\section{Introduction} \label{introduction}

Recently, vehicle-mounted mobile mapping systems (MMSs) have emerged as the primary spatial imaging system for capturing high-resolution maps of urban environments. Most commercially-available MMSs combine georeferencing technology with precise, high-speed, long-range laser scanning and high-resolution imaging sensors. In \cite{elhashash2022review}, a comprehensive analysis of contemporary MMSs was presented. With the most recent MMS technology, it is now possible to collect enormous quantities of precise, location-based data for various purposes. This includes creating detailed maps suitable for autonomous vehicles and the use of these maps to generate large-scale 3D representations of entire cities \cite{al2019road}.
\begin{figure} [t]
	\centering
	\includegraphics[width=85mm]{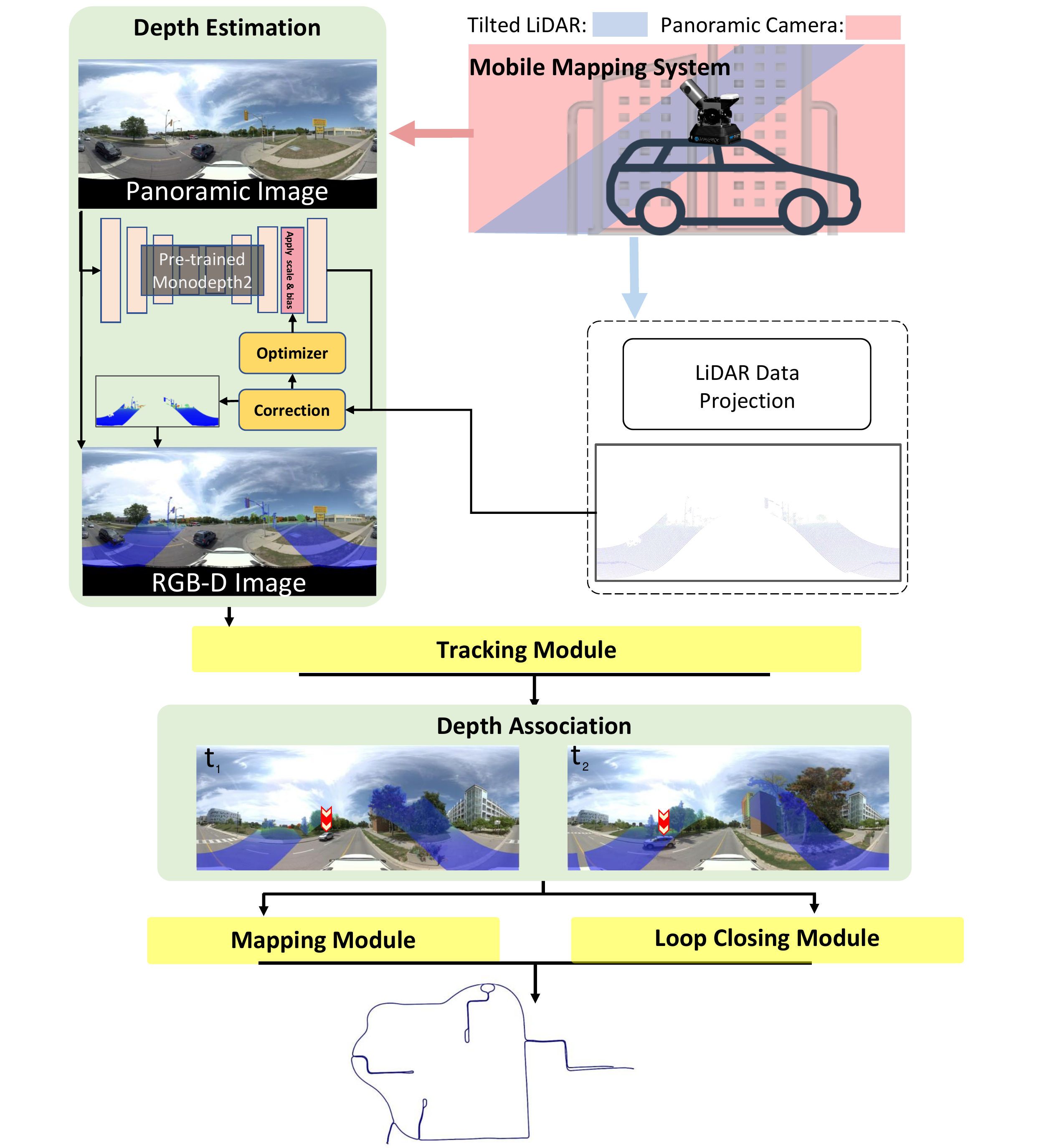}
	\caption{
	A schematic workflow designed for HDPV-SLAM using MMS's panoramic camera and tilted LiDAR scanner. HDPV-SLAM proposes new modules including depth estimation and depth association (shown in green blocks).
	}
	\label{fig1_mms_our_interest}
\end{figure}

However, the georeferencing capability of MMS significantly relies on advanced GNSS/IMU technology. The primary purposes of spatial imaging sensors such as cameras and LiDAR are to generate photorealistic textures and metrically-scaled depths. In GNSS-unfavorable environments, such as urban canyons and tunnels, MMS georeferences spatial data with limitations. In addition, we have recently observed a high demand for compact and inexpensive MMS designs to reduce system complexity, including using GNSS/IMU less frequently or not at all \cite{elhashash2022review}.

This investigation has two primary objectives. The first objective is to incorporate a mapping-capable MMS with a SLAM system. The MMS we utilized was equipped with a panoramic camera along with a tilted LiDAR, which is usually employed for road inspection or mapping. The second objective is to construct a metrically-scaled, accurate SLAM system that could operate in GPS-denied environments without the costly IMU sensor. SLAM systems are constrained by GPS limitations on use in such environments, and while high-end IMU sensors can circumvent this obstacle, they may be too costly for low-budget users. To address this issue, we conducted tests to ascertain the performance precision of inexpensive visual sensors in a boundary test scenario.

Our foundational work, RPV-SLAM \cite{kang2021rpv}, traverses the path to attaining the stated objectives but has certain limitations. RPV-SLAM is a keyframe-based RGB-D SLAM that generates metrically-scaled trajectories with comparatively small errors using only LiDAR along with a panoramic camera. According to the method, the depth channel of its RGB-D images is supplied by a LiDAR sensor. They show that the LiDAR field-of-view is sufficient enough to give us approximately correct metrically-scaled results. However, there are two restrictions in RPV-SLAM. First, it uses a simplistic bi-interpolation for LiDAR densification which disregards the scene's context and semantics that could be inferred from the corresponding imagery. Second, as a keyframe-based SLAM, RPV-SLAM generates 3D map points only with the keyframes using two independent modes of depth association, one powered by visual-feature-based triangulation and the other by LiDAR. Due to the significant disparity between the horizontal field of view provided by the panoramic camera and the tilted LiDAR scanner, multi-modal depth generation is required. Thus, the quality of map points generated by multi-modal depth generation varies depending on the ad hoc positioning of extracted features in the keyframes. 

To address these issues, we propose HDPV-SLAM system, which employs two novel modules to enhance the precision of MMS's pose estimation. In HDPV-SLAM pipeline, we designed a deep-learning (DL)-based depth estimation module to resolve the limitations caused by the naive bi-interpolation for densifying the depth from the MMS's panoramic camera and tilted LiDAR sensor. DL-based depth estimation has shown a high level of proficiency in providing semantic-centric depth information to enhance the visual SLAM's performance \cite{tateno2017cnn}. However, the lack of real, depth-annotated datasets and the constantly-changing environment of SLAM systems, such as different lighting and weather conditions, hinder the generalization of deep depth estimation networks. This results in a subpar performance in practice and during the inference phase for complicated or unseen inputs. Using supervised and unsupervised approaches \cite{zhao2022unsupervised, yen20223d}, various adaptation methods attempt to alleviate this problem. For instance, DARS \cite{alizadeh_naeini_adaptive_2022} is designed to use a pre-trained depth estimation network in conjunction with available sparse LiDAR points during inference and to adapt the network to each new input. Although DARS can provide us with a more precise and dense depth estimation, it was designed for perspective images, whereas the images in our problem have a panoramic geometry. In this study, we propose Panoramic Double-stage Adaptive Refinement Scheme (PanoDARS), an extension of DARS for panoramic images that generates DL-based depth estimation results approximated to LiDAR depth during the inference phase. 

To circumvent the second limitation of RPV-SLAM, we propose a depth association module that combines keyframe and non-keyframe frame (or non-keyframe for simplicity) depth information. Because our tilted LiDAR sensor has a small overlap region with the panoramic camera, two distinct depth generation procedures are carried out. First, generate 3D map points with keyframes using DL-based depth estimation if visual features are found in the overlapping region. Second, perform visual feature-based triangulation outside the overlapping region. Two distinct sensing mechanisms generate map points with varying degrees of precision. In our proposed depth association module, we cross-validate the accuracy of the single modal map point creation with the dual modalities and, if necessary, update the position of the associated map point with depth information from non-keyframes. This cross-validation task is accomplished by extending the depth association's capacity with non-keyframes, allowing visual feature-only map points to be tracked and compared with LiDAR-only map points located in an overlapping region. This cross-validation of multi-modal depth estimation results can enhance the positional accuracy of map points and maximize the use of available depth information from the LiDAR sensor.

In this study, we consider HDPV-SLAM’s major contributions as follows:
\begin{itemize}
\item We propose a novel DL-based depth estimation module for iteratively densifying sparse LiDAR depth using panoramic images and LiDAR point clouds to improve the quality of depth estimation on visual features.
\item We propose a novel hybrid depth association module that optimally combines depth information driven by triangulating visual features and LiDAR depth to address a problem caused by the lack of substantial spatial overlap between the panoramic camera and the tilted LiDAR sensor.
\end{itemize}

We review related works in Section \ref{section_related_works}. The proposed HDPV-SLAM system is presented in Section \ref{methodology}, and experimental results are discussed in Section \ref{experiments}.
Finally, we present the concluding remarks on our work in Section \ref{conclusion}.

\section{Related Works}		\label{section_related_works}
\subsection{Visual SLAM}
In recent years visual SLAM and odometry have had various remarkable works, including ORB-SLAM \cite{mur2017orb}, LSD-SLAM \cite{engel2015large}, DSO \cite{engel2017direct}, and SVO \cite{forster2014svo}. The SOTA visual SLAM can be classified into filter-based SLAM and keyframe-based SLAM. MonoSLAM \cite{davison2007monoslam} is one of the filter-based methods that use the extended Kalman filter (EKF). In contrast, keyframe-based visual SLAM only utilizes selected frames known as keyframes for mapping instead of processing all frames, while non-keyframes are utilized for tracking purposes. Keyframes are also used in the loop closing module which detects the revisited keyframes and improves the performance of bundle adjustment optimization for large-scale environments. \cite{YOUNES201767} presents a literature review on keyframe-based SLAM systems.

\subsection{RGB-D SLAM}
RGB-D SLAMs have gained considerable attention in the literature due to their high accuracy and are being widely used in real-world applications. These methods are a type of visual SLAM that have been proposed for RGB-D cameras and their trend mostly follows the monocular methods. Just like the monocular visual SLAMs, RGB-D SLAMs can be categorized into feature-based and direct methods. Feature-based methods like \cite{mur2017orb, LI202210} extract features from the image and use the geometry of the reconstructed scene for the camera pose estimation. Although these methods fail in certain environments where their feature extractor fails, they are computationally efficient compared with the direct methods. However, direct methods like \cite{newcombe2011dtam, whelan2015elasticfusion}, on the other hand, utilize the whole RGB-D image to minimize the photometric error between the predicted image and the observed image. But the big downside of these methods is that they are more computationally expensive.

\subsection{Metrically-scaled SLAM}
To solve the scale ambiguity problem of visual SLAM results, a diverse range of methods have been suggested so far, and it is quite an old topic in the literature. Some propose the usage of auxiliary sensors. For example, \cite{shi2012gps} uses GPS data, and \cite{ji2020panoramic} uses ground control points; both have limited use cases. In contrast, PIW-SLAM \cite{jiang2021panoramic} generates metrically-scaled results with the help of IMU and wheel encoder, which does not limit the SLAM for real-world applications. Furthermore, the works of \cite{kerl2013robust, huai2015real} use RGB-D cameras like Kinect providing aligned dense depth data for monocular images.

In the context of solving the ambiguity problem, the utilization of RGB-D-like images has been efforts, too. \cite{tiwari2020pseudo} utilizes a deep neural network to predict a depth image for each frame and have it as the $D$ channel in RGB-D images for the input of their RGB-D SLAM. Similar to the previous work, \cite{scalerecover2017xiaochuan} proposes the usage of deep convolutional neural fields for depth estimation. RPV-SLAM \cite{kang2021rpv} uses a sparse LiDAR sensor and using the bi-interpolation of the LiDAR points, forms the D channel of the input RGB-D image along with the panoramic RGB image. Although usage of a LiDAR sensor with a pinhole camera in the SLAM community has been addressed and has proven its strengths \cite{chou2021efficient}, utilizing a LiDAR sensor with a panoramic camera has been an undervalued topic, and to the best of our knowledge RPV-SLAM is the only one.

\subsection{Depth Estimation}
Depth estimation is called to the process of estimating depth for input RGB images. Recently, It has become a hot topic due to the success of deep learning methods \cite{bhat2021adabins, ranftl2021vision} and its various applications ranging from 3D reconstruction \cite{zhuang2022acdnet} to autonomous vehicles \cite{fonder2021m4depth}. RGB-D SLAM often suffers from sparse depth measurements and depth estimation suffers from scale ambiguity and lack of ground truth data, so these two tasks can contribute to each other \cite{jin2021mono}. In this sense, a number of works have tried improving the performance of depth estimation \cite{sartipi2020deep}, SLAM \cite{tateno2017cnn} or even both \cite{loo2021online} through a complementary design. 

Unlike the tremendous research on perspective depth estimation, considerably fewer studies have been dedicated to panoramic depth estimation. One main reason for that is the lack of appropriate panoramic datasets for depth estimation \cite{yun2022improving}. Therefore, some research such as \cite{de2018eliminating} suggest training existent networks on synthetic panoramic images. Another approach is domain adaptation from perspective to panoramic. For instance, \cite{tateno2018distortion} takes the available pre-trained networks on perspective images and replaces the standard convolutions with a distortion-aware version of convolutions to infer on panoramic images. OmniFusion \cite{li2022omnifusion} also transforms panoramic images into perspective images, then estimates the depth and eventually merges them to reconstruct the panoramic depth. As another approach, \cite{yun2022improving} advocates self-supervised learning to eliminate the need for panoramic depth data during training. Furthermore, it is worth mentioning that, among all the above approaches, adaptation of perspective models to the panoramic domain apparently has gained increasing attention as a practical solution in other tasks as well \cite{zhang2022bending, zhang2021transfer}.

\begin{figure*}[ht]
\centering
     \includegraphics[width=1.0\textwidth]{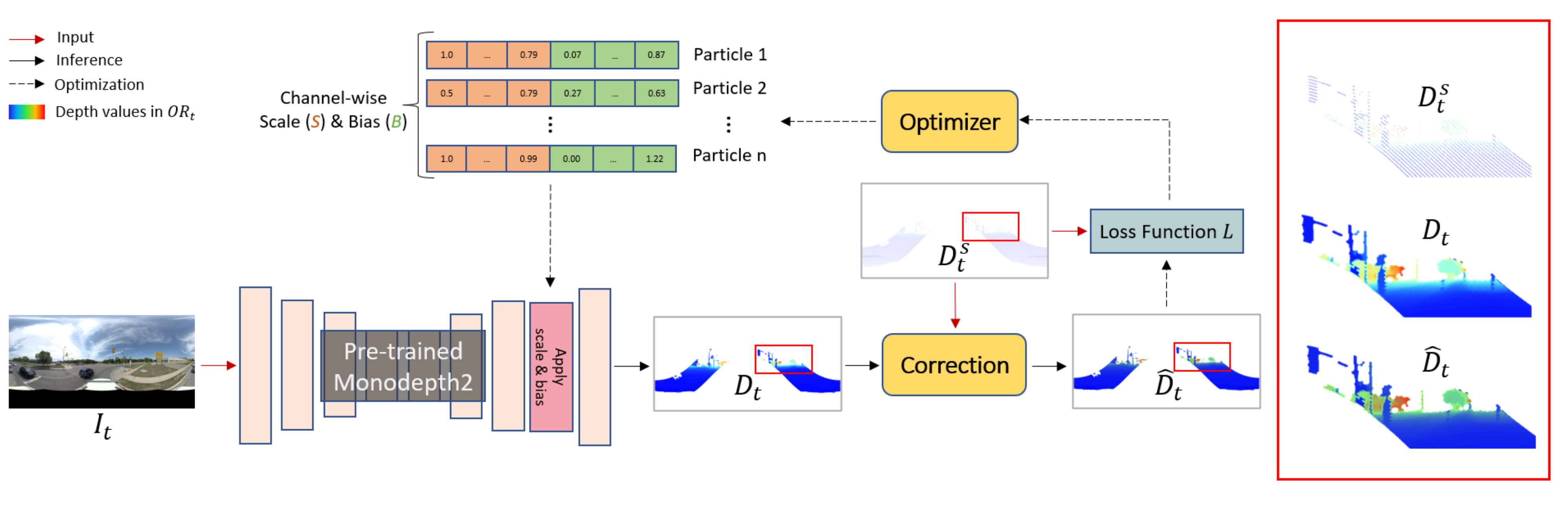}
      \caption{A schematic architecture designed for the depth estimation module, called PanoDARS. PanoDARS uses panoramic imagery as a primary source for predicting the depths to generate RGB-D images, while LiDAR data is used as supplementary information to correct the predicted depth. A pre-trained network predicts a depth image $\textbf{D}_t$ using panoramic imagery as primary source. Given a sparse depth $\textbf{D}^s_t$, the predicted $\textbf{D}_t$ is adjusted to the acquired LiDAR point clouds for generating a corrected depth $\hat{\textbf{D}}_t$. Finally, a refined dense depth is generated by optimizing embedded feature space to minimize the loss between $\textbf{D}^s_t$ and $\hat{\textbf{D}}_t$ in an iterative manner during the inference phase.}
       \label{depth_estimation_figure}
\end{figure*}

\section{Hybrid Depth-augmented Panoramic \\Visual SLAM} \label{methodology}
As shown in Figure \ref{depth_estimation_figure}, HDPV-SLAM includes modules for depth estimation, depth association, tracking, mapping, and loop closing modules. This section describes each module in detail.

\subsection{SLAM Input} \label{lidar_projection}
In the first stage, an RGB-D image is created by combining a panoramic RGB image and its corresponding LiDAR point cloud. The image channel $D$ is created by projecting LiDAR data onto the image plane. Using calibration parameters between the camera and the LiDAR sensor, this task is completed in accordance with the prior work \cite{kang2021rpv}.

Then, for each RGB-D image in the capturing order, frame $F_t$ is created, where  $t \in \{1,2,3, ..., n\}$ and $n$ is the total number of RGB-D images. $\textbf{F}$ is the array of all frames. Each $F_t \in \textbf{F}$ contains RGB image $\textbf{I}_t\in\mathbb{R}^{w\times h\times 3}$ of the width $w$ and the height $h$ and a corresponding projected sparse LiDAR $\textbf{D}^s_t\in\mathbb{R}^{w\times h}$. In Section \ref{depth_estimation}, we describe a method to densify $\textbf{D}^s_t$ and get $\hat{\textbf{D}}_t\in\mathbb{R}^{w\times h}$. Also, a frame $F_t$ contains $O_t$ which is the array of the ORB features extracted from $\textbf{I}_t$ using the ORB feature extractor. For each $o^i_t \in O_t$ we have $o^i_t \in \{(u, v)| u, v \in \mathbb{W}, 0 \leq u < w, 0 \leq v < h\}$where $i \in \{1, 2, 3, ..., m_t\}$ and $m_t$ is the total number of ORB features of $I_t$.

The output of the SLAM system is the array of all the camera positions $J$ and the set of all the map points $\rho$. Each $j_t \in \mathbb{R}^3\times SO(3)$ in $J$ is the position and the orientation of the camera at time $t$ in world coordinate, and each $p \in \mathbb{R}^3$ in $\rho$ shows the position of a map point in the world coordinate. We define subsets of $\rho$, for each $t$ as we create map points using $F_t$. The relation between $\rho_t$ for different $t$ is $\{\} = \rho_0 \subset \rho_1 \subseteq \rho_2 \subseteq ... \rho_{n-1} \subseteq \rho_n = \rho$.  Each $\rho_t \subset \rho$ is updated with the new map points created from $F_t$ each time a new frame $F_t$ comes in. And when $t=n$, we have $\rho_t = \rho$

\subsection{Depth Estimation Module} \label{depth_estimation}
The depth estimation module aims to produce a dense depth map $\hat{\textbf{D}}_t$ by taking an input RGB image $\textbf{I}_t$ at the time $t$ and a corresponding sparse LiDAR $\textbf{D}^s_t$. This module enables increasing the number of ORB visual features augmented with depth in the overlapping region ($OR_t$) between the panoramic imagery and the LiDAR point clouds. To do this, a new version of DARS \cite{alizadeh_naeini_adaptive_2022} for panoramic depth estimation, called PanoDARS, is proposed (see Figure \ref{depth_estimation_figure}). DARS \cite{alizadeh_naeini_adaptive_2022} was originally designed for adaptive depth refinement on perspective images, while PanoDARS has to adapt the models pre-trained on perspective images to panoramic ones. However, depth estimation on panoramic images presents some obstacles. First, because there is no publicly accessible standard dataset for panoramic depth estimation, supervised models cannot be used. Second, self-supervised methods must be improved by non-metrically scaled estimations and are notably difficult for panoramic geometry. Thus, the suggested solution is to adapt a pre-trained depth estimation baseline over perspective images to our panoramic dataset.

\begin{figure*}[ht]
\centering
     \includegraphics[width=1.0\textwidth]{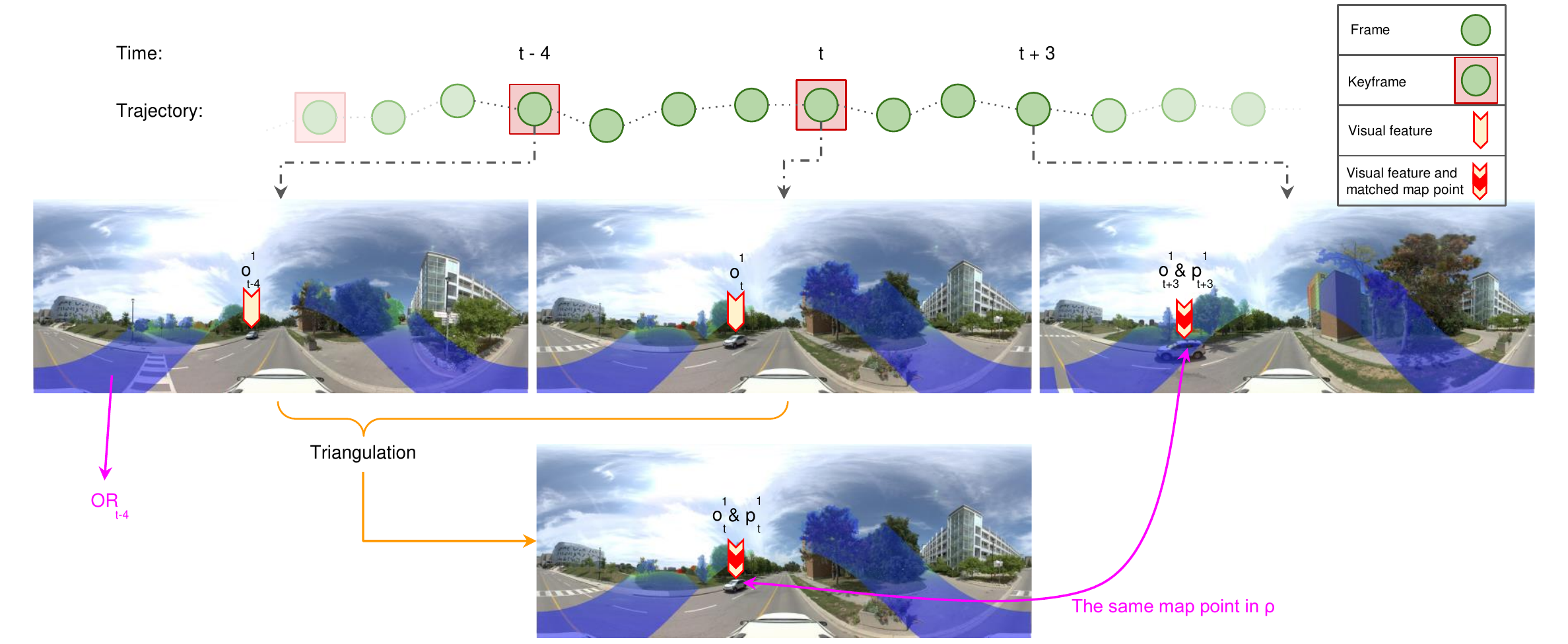}
      \caption{A map point $p^1_{t}$ is created at time $t$ by triangulating visual features of times $t-4$ and $t$. This triangulation is because these two visual features are not in $OR_{t-4}$ and $OR_t$. In $F_{t+3}$, $o^1_{t+3}$ points to the middle of the parked car's rooftop and matches with the previously created $p^1_{t}$ at time $t$ which is now called $p^1_{t+3}$. At time $t+3$, $o^1_{t+3}$ is inside $OR_{t+3}$, which means the depth data is available. This enables the system to estimate the map points' 3D position more precisely through the depth coming from the depth estimation method.}
       \label{depth_association_module_figure}
\end{figure*}

PanoDARS consists of two stages, correction and optimization. In \cite{alizadeh_naeini_adaptive_2022}, the depth maps are divided into three horizontal slices, with the correction values in each slice being independently calculated. Since LiDAR sparsity patterns differ between panoramic and perspective geometries, the proposed method omits slicing during the correction phase. Moreover, PanoDARS only estimates depth in $I_t$ regions where sparse depth is available nearby ($OR_t$, as shown in the rainbow-colored image on the left of Figure \ref{depth_association_module_figure}).

In the first stage, a correction value $\delta d^s_t \in \Delta \textbf{D}^s_t$ between each valid pixel in the sparse depth map $ d^s_t \in  \textbf{D}^s_t$ and its correspondence $d_t \in \textbf{D}_t$ is calculated using $\delta d^s_t=d_t-d^s_t$.

Then, an interpolation function $Q: \mathbb{R}^2 \mapsto \mathbb{R}$ based on Delaunay triangulation \cite{amidror_scattered_2002} is leveraged to obtain a dense correction map $\Delta \textbf{D}_t = Q(\Delta \textbf{D}^s_t)$. Finally, the corrected depth map $\hat{\textbf{D}}_t = \textbf{D}_t + \Delta \textbf{D}_t$ is calculated. $\hat{\textbf{D}}_t$ is a sufficiently accurate initial value for the second stage.

In the second stage, while the pre-trained weights are fixed, some learnable auxiliary parameters are applied to intermediate features in the baseline. By optimizing those parameters, the predicted depth is refined. Therefore, the overall performance of PanoDARS is as follows.

Given an input RGB image $\textbf{I}_t\in\mathbb{R}^{w\times h\times 3}$, PanoDARS splits the pre-trained baseline $M:\mathbb{R}^{w \times h \times 3}\mapsto\mathbb{R}^{w \times h}$ into a body $G:\mathbb{R}^{w \times h \times 3}\mapsto\mathbb{R}^{a \times b \times c}$ and a head $H: \mathbb{R}^{a \times b \times c}\mapsto\mathbb{R}^{w \times h}$, where $a$, $b$, and $c$ are respectively width, height, and number of channels of the intermediate feature set $G(\textbf{I}_t)\in\mathbb{R}^{a\times b\times c}$. The auxiliary parameters, scales $\textbf{S} \in \mathbb{R}^c$ and biases $\textbf{B} \in \mathbb{R}^c$ are applied on $G(I_t)$, and the depth $\textbf{D}_t = H(\textbf{S}\bigotimes G(\textbf{I}_t)\bigoplus \textbf{B})$ is predicted, where $\bigotimes$ and $\bigoplus$ represent channel-wise multiplication and addition. Afterwards, the correction module $C: \mathbb{R}^{w \times h}\mapsto\mathbb{R}^{w \times h}$ carries out the first refinement stage on $\textbf{D}_t$ and returns $\hat{\textbf{D}}_t= C(\textbf{D}_t, \textbf{D}^s_t)$.

The auxiliary parameters $\textbf{X} \in \mathbb{R}^{2c}$, i.e., concatenated channel-wise scales ($\textbf{S}$) and biases ($\textbf{B}$), are learnable. Therefore, the following optimization problem can be formulated as $L(\hat{\textbf{D}}_t(\textbf{I}_t,\textbf{X}+\Delta \textbf{X}), \textbf{D}^s_t) \to \min\limits_{\Delta \textbf{X}}.$

where $\Delta \textbf{X}$ is the corrections applied on the parameters and $L$ is the cost function given to particle swarm optimizer (PSO) \cite{kennedy_particle_1995}. Hence, the second stage of refinement is conducted by PSO.

\subsection{Tracking Module} \label{tracking}
The tracking module establishes relationships between extracted visual features and existing map points. The tracking module accepts an input of $F_t$ for each $t$ in sequential order at the start of a SLAM run. The matching process within the tracking module will locate matches between $\rho_{t-1}$ and generate an array of matched map points $P_t$ and $O_t$. $P_t$ has the same size as $O_t$ (which is $m_t$) and can also contain $Null$ values at certain indices. For each matched pair of $o^i_t$ and a map point from $\rho_{t-1}$, $P_t$ contains a $p^i_t$ pointing to the map point. $p^i_t = Null$ if $o^i_t$ does not match any map point from $\rho_{t-1}$. The tracking module can estimate $j_t$ based on these local matches. The system then designates a frame $F_t$ as a keyframe when it determines that the tracking will perform poorly in the future with estimates of $j_t$ based on a limited number of map points.

\subsection{Mapping Module} \label{mapping}
When the tracking module decides $F_t$ to be a keyframe, the mapping module creates new map points using $F_t$ and puts them together with $\rho_{t-1}$ to save all of them in $\rho_t$. 
The module attempts to create a map point for each $o^i_t$ that its corresponding $p^i_t$ is $Null$. If $o^i_t \in OR_t$, it creates a map point using $o^i_t$, $j_t$, and $\hat{\textbf{D}}_t(o^i_t)$. If $o^i_t \notin OR_t$, it will triangulate $F_t$ with the nearest neighboring keyframe $F_s$ to estimate the depth for $o^i_t$ and create the map point with the same set of information. If triangulation is not successful, the module will not create a map point for $o^i_t$. The mapping module is also responsible for local bundle adjustment, which is accomplished by following \cite{kang2021rpv}.

\subsection{Depth Association Module} \label{depth_association}

We propose a novel hybrid depth association module for associating optimal depth information with visual features in $OR_t$.  After the tracking module has tracked all visible map points in the current frame $F_t$ and estimated $j_t$, this depth association module is triggered. The proposed module is executed even if $F_t$ is not a keyframe for augmenting $O_t$ with $\hat{\textbf{D}}_t$, unlike conventional methods. A key motivation for using non-keyframes in the depth association is to cross-validate the accuracy of tracked map points and update the map points by conducting the depth association optimally. As previously described, a map point can be generated using either the direct assignment of LiDAR-driven depth to a visual feature or the visual feature-based triangulation; it depends on the position of the corresponding feature in the image plane (i.e., being inside $OR_t$ or out of it). Consequently, the depth value associated with a tracked map point may vary. The new method resolves this inconsistency issue by executing an optimal depth association task following established policies.

Algorithm \ref{alg:cdm} shows the suggested hybrid depth association process. It inputs $F_t$ and the tracked array of map points $P_t$. For each $i$, if the matching module has not assigned any map points to $o^i_t$ or if $o^i_t$ is out of $OR_t$, it skips the $i$ and goes for the next member of $O_t$. If $o^i_t$ passes these conditions, then a ${}^{new}p^i_t \in \mathbb{R}^{3}$ will be created using $o^i_t$, its depth value ($\hat{\textbf{D}}_t(o^i_t)$), and its depth value and the position and orientation of the camera at time $t$ ($j_t$). 
But, sometimes the tracking module can make mistakes in assigning ORB visual features to the corresponding map points. So we have defined a rejection threshold $\theta$ (in meters) to prevent big mistakes and their effect on the whole trajectory. If the distance between the current position of $p^i_t$ and the newly calculated ${}^{new}p^i_t$ is bigger than $\theta$, we will not change anything.
If not, then we have defined two policies here:
\begin{itemize}
\item If $p^i_t$ has been created using triangulation and not modified using any depth map later, the algorithm will change the location of $p^i_t$ to ${}^{new}p^i_t$.
\item Or, if the depth value of the visual feature that previously has modified (or created) the position of $p^i_t$ ($\hat{\textbf{D}}_p(o^i_p)$) is larger than the current depth ($\hat{\textbf{D}}_t(o^i_t)$), again the algorithm will change the location of $p^i_t$ to ${}^{new}p^i_t$.
\end{itemize}

Figure \ref{depth_association_module_figure} shows an example of this algorithm for the first policy with a simple scenario where we only have one ORB feature in each frame. And the second policy helps with the precise estimation of the position of the map points. Each time a map point is observed with a smaller depth value, the algorithm recalculates its position based on it.

\begin{algorithm}
\footnotesize
\caption{Depth Association Module Algorithm}\label{alg:cdm} 
\textbf{Input: Current frame $F_t$ and the tracked array of map points $P_t$}
\begin{algorithmic}[1]

\State $\theta \gets ${The rejection threshold}

\For{Each $p^i_t$ in   $P_t$}
    \If{$p^i_t = Null$ or $o^i_t \notin OR_t$}
        \State $continue$
    \EndIf
    \State $F_p \gets ${Last keyframe who created or modified $p^i_t$}
    \State ${}^{new}p^i_t \gets ${Estimated 3D position using $o^i_t$, $\hat{\textbf{D}}_t(o^i_t)$, and $j_t$}
    \If{distance($p^i_t$, ${}^{new}p^i_t$) $\textless \theta$}
        \If{($p^i_t$ created using triangulation and not modified after that) or ($\hat{\textbf{D}}_t(o^i_t) \textless \hat{\textbf{D}}_p(o^i_p)$)}
            \State $p^i_t$'s position $\gets {}^{new}p^i_t$
            \State Update $p^i_t$ in $\rho_t$ and all the previous subsets
        \EndIf
    \EndIf
\EndFor
\end{algorithmic}
\end{algorithm}

\subsection{Loop Closing Module}\label{loop_closure}
Finally, after the loop closing module detects a loop, it performs the global bundle adjustment on the trajectory inside the loop. The global bundle adjustment performs a pose-graph optimization on $j_t$ of the camera in each $F_t \in$ \{the subset of keyframes in the loop\}. And also it optimizes all of the map points created in those keyframes. For the pose-graph optimization, we have used a similarity transformation.

\section {Experimental Results and Discussion} \label{experiments}
This section comprises experimental analyses that examine the individual and combined performance of two modules through an ablation study. Additionally, we assess the performance of the most effective method with the combination of two modules identified in the ablation study on the entire dataset. We used York University and Teledyne Optech (YUTO) MMS dataset. The dataset has been acquired by Teledyne Optech's Maverick MMS with four sequences in various outdoor environments with an 18.95 km long road in York University's Keele Campus to test different urban situations. The characteristic details of the dataset and sensor details can be found in the reference \cite{kang2021rpv}. All the experiments have been conducted offline and post-data acquisition.

\begin{table*}[t]
\label{table_result_ablation}
\caption{\label{tab:table_result_ablation}SLAM trajectory results for ablation study on Sequence B.}
\begin{center}
\begin{tabular}{|c|c|c||c|c|c|c|}
\hline
SLAM & 
Densification method &
$\theta$ (m) &
ATE (m) &
RTE ($ \% $) &  
RRE ($ ^ \circ $/m)
\\ \hline
\begin{tabular}[c]{@{}c@{}} \\[-1em] Cartographer 									\end{tabular} & 
\begin{tabular}[c]{@{}c@{}} N/A		\end{tabular} & 
\begin{tabular}[c]{@{}c@{}} N/A		\end{tabular} & 
\begin{tabular}[c]{@{}c@{}} 142.97		\end{tabular} & 
\begin{tabular}[c]{@{}c@{}} 16.57		\end{tabular} & 
\begin{tabular}[c]{@{}c@{}}  0.0093 	\end{tabular}  
\\ \hline
\begin{tabular}[c]{@{}c@{}} \\[-1em] RPV-SLAM			\end{tabular} &
\begin{tabular}[c]{@{}c@{}} bi-interpolation	\end{tabular} & 
\begin{tabular}[c]{@{}c@{}} N/A		\end{tabular} & 
\begin{tabular}[c]{@{}c@{}} 12.91		\end{tabular} & 
\begin{tabular}[c]{@{}c@{}} 1.51		\end{tabular} & 
\begin{tabular}[c]{@{}c@{}} {\textbf{0.0009}}			\end{tabular}  
\\ \hline
\begin{tabular}[c]{@{}c@{}} \\[-1em] Proposed method		\end{tabular} &
\begin{tabular}[c]{@{}c@{}} depth estimation module		\end{tabular} & 
\begin{tabular}[c]{@{}c@{}} N/A		\end{tabular} & 
\begin{tabular}[c]{@{}c@{}} 12.95		\end{tabular} & 
\begin{tabular}[c]{@{}c@{}} 1.64		\end{tabular} & 
\begin{tabular}[c]{@{}c@{}} {0.0010}			\end{tabular}  
\\ \hline \hline
\begin{tabular}[c]{@{}c@{}} \\[-1em]Proposed method \end{tabular} &
\begin{tabular}[c]{@{}c@{}} bi-interpolation		\end{tabular} & 
\begin{tabular}[c]{@{}c@{}} 1		\end{tabular} & 
\begin{tabular}[c]{@{}c@{}} 11.99		\end{tabular} & 
\begin{tabular}[c]{@{}c@{}} {1.62}		\end{tabular} & 
\begin{tabular}[c]{@{}c@{}} {\textbf{0.0009}}			\end{tabular}  
\\ \hline
\begin{tabular}[c]{@{}c@{}} \\[-1em]Proposed method			\end{tabular} &
\begin{tabular}[c]{@{}c@{}} bi-interpolation		\end{tabular} & 
\begin{tabular}[c]{@{}c@{}} 2		\end{tabular} & 
\begin{tabular}[c]{@{}c@{}} 9.93		\end{tabular} & 
\begin{tabular}[c]{@{}c@{}} 1.43		\end{tabular} & 
\begin{tabular}[c]{@{}c@{}} {0.0010}			\end{tabular}  
\\ \hline
\begin{tabular}[c]{@{}c@{}} \\[-1em]Proposed method \end{tabular} &
\begin{tabular}[c]{@{}c@{}} bi-interpolation		\end{tabular} & 
\begin{tabular}[c]{@{}c@{}} 3		\end{tabular} & 
\begin{tabular}[c]{@{}c@{}} 10.19		\end{tabular} & 
\begin{tabular}[c]{@{}c@{}} 1.56		\end{tabular} & 
\begin{tabular}[c]{@{}c@{}} {\textbf{0.0009}}			\end{tabular}  
\\ \hline
\begin{tabular}[c]{@{}c@{}} \\[-1em]Proposed method \end{tabular} &
\begin{tabular}[c]{@{}c@{}} bi-interpolation		\end{tabular} & 
\begin{tabular}[c]{@{}c@{}} 4		\end{tabular} & 
\begin{tabular}[c]{@{}c@{}} 13.01		\end{tabular} & 
\begin{tabular}[c]{@{}c@{}} 1.63		\end{tabular} & 
\begin{tabular}[c]{@{}c@{}} {0.0010}			\end{tabular}  
\\ \hline
\begin{tabular}[c]{@{}c@{}} \\[-1em]Proposed method\end{tabular} &
\begin{tabular}[c]{@{}c@{}} bi-interpolation		\end{tabular} & 
\begin{tabular}[c]{@{}c@{}} 5		\end{tabular} & 
\begin{tabular}[c]{@{}c@{}} 13.70		\end{tabular} & 
\begin{tabular}[c]{@{}c@{}} 1.74		\end{tabular} & 
\begin{tabular}[c]{@{}c@{}} {0.0010}			\end{tabular}  
\\ \hline \hline

\begin{tabular}[c]{@{}c@{}} \\[-1em]Proposed method \end{tabular} &
\begin{tabular}[c]{@{}c@{}} depth estimation module		\end{tabular} & 
\begin{tabular}[c]{@{}c@{}} 1		\end{tabular} & 
\begin{tabular}[c]{@{}c@{}} 11.28		\end{tabular} & 
\begin{tabular}[c]{@{}c@{}} 1.54		\end{tabular} & 
\begin{tabular}[c]{@{}c@{}} {\textbf{0.0009}}			\end{tabular}  
\\ \hline
\begin{tabular}[c]{@{}c@{}} \\[-1em]Proposed method			\end{tabular} &
\begin{tabular}[c]{@{}c@{}} depth estimation module		\end{tabular} & 
\begin{tabular}[c]{@{}c@{}} 2		\end{tabular} & 
\begin{tabular}[c]{@{}c@{}} {\textbf{9.58}}		\end{tabular} & 
\begin{tabular}[c]{@{}c@{}} {\textbf{1.23}}		\end{tabular} & 
\begin{tabular}[c]{@{}c@{}} {0.0011}			\end{tabular}  
\\ \hline
\begin{tabular}[c]{@{}c@{}} \\[-1em]Proposed method \end{tabular} &
\begin{tabular}[c]{@{}c@{}} depth estimation module		\end{tabular} & 
\begin{tabular}[c]{@{}c@{}} 3		\end{tabular} & 
\begin{tabular}[c]{@{}c@{}} 11.96		\end{tabular} & 
\begin{tabular}[c]{@{}c@{}} 1.65		\end{tabular} & 
\begin{tabular}[c]{@{}c@{}} {\textbf{0.0009}}			\end{tabular}  
\\ \hline
\begin{tabular}[c]{@{}c@{}} \\[-1em]Proposed method \end{tabular} &
\begin{tabular}[c]{@{}c@{}} depth estimation module		\end{tabular} & 
\begin{tabular}[c]{@{}c@{}} 4		\end{tabular} & 
\begin{tabular}[c]{@{}c@{}} 12.24		\end{tabular} & 
\begin{tabular}[c]{@{}c@{}} 1.52		\end{tabular} & 
\begin{tabular}[c]{@{}c@{}} {0.0010}			\end{tabular}  
\\ \hline
\begin{tabular}[c]{@{}c@{}} \\[-1em]Proposed method\end{tabular} &
\begin{tabular}[c]{@{}c@{}} depth estimation module		\end{tabular} & 
\begin{tabular}[c]{@{}c@{}} 5		\end{tabular} & 
\begin{tabular}[c]{@{}c@{}} 14.16		\end{tabular} & 
\begin{tabular}[c]{@{}c@{}} 1.95		\end{tabular} & 
\begin{tabular}[c]{@{}c@{}} {0.0010}			\end{tabular}  
\\ \hline

\end{tabular}

\end{center}
\end{table*}

\begin{table*}[t]
\label{table_result_all}
\caption{\label{tab:table_result_all}SLAM trajectory results for all the sequences. Settings of HDPV-SLAM is based on Table \ref{tab:table_result_ablation}.}
\begin{center}
\begin{tabular}{|c||c|c|c|c||c|c|c|c|}
\hline
SLAM &
Sequence & 
ATE (m) &
RTE ($ \% $) & 
RRE ($ ^ \circ $/m) &
Sequence & 
ATE (m) &
RTE ($ \% $) &   		
RRE ($ ^ \circ $/m)

\\ \hline  \hline
\begin{tabular}[c]{@{}c@{}} \\[-1em] Cartographer \end{tabular} & 
\begin{tabular}[c]{@{}c@{}} A		\end{tabular} & 
\begin{tabular}[c]{@{}c@{}} 4.15		\end{tabular} & 
\begin{tabular}[c]{@{}c@{}} 6.79		\end{tabular} & 
\begin{tabular}[c]{@{}c@{}}  0.0507 	\end{tabular}  &
\begin{tabular}[c]{@{}c@{}} C		\end{tabular} & 
\begin{tabular}[c]{@{}c@{}} 180.95		\end{tabular} & 
\begin{tabular}[c]{@{}c@{}} 5.78		\end{tabular} & 
\begin{tabular}[c]{@{}c@{}}  0.0133 	\end{tabular}  

\\ \hline
\begin{tabular}[c]{@{}c@{}} \\[-1em] RPV-SLAM \end{tabular} &
\begin{tabular}[c]{@{}c@{}} A	\end{tabular} & 
\begin{tabular}[c]{@{}c@{}} 1.62		\end{tabular} & 
\begin{tabular}[c]{@{}c@{}} \textbf{3.25}		\end{tabular} &
\begin{tabular}[c]{@{}c@{}} {0.0268}			\end{tabular} &

\begin{tabular}[c]{@{}c@{}} C	\end{tabular} & 
\begin{tabular}[c]{@{}c@{}} 30.66		\end{tabular} & 
\begin{tabular}[c]{@{}c@{}} {2.82}		\end{tabular} & 
\begin{tabular}[c]{@{}c@{}} {0.0042}			\end{tabular} 
\\ \hline
\begin{tabular}[c]{@{}c@{}} \\[-1em] Proposed method \end{tabular} &
\begin{tabular}[c]{@{}c@{}} A		\end{tabular} & 
\begin{tabular}[c]{@{}c@{}} {\textbf{1.40}}	\end{tabular} & 
\begin{tabular}[c]{@{}c@{}} 3.59		\end{tabular} & 
\begin{tabular}[c]{@{}c@{}} \textbf{0.0041}		\end{tabular} &

\begin{tabular}[c]{@{}c@{}} C		\end{tabular} & 
\begin{tabular}[c]{@{}c@{}} \textbf{11.93}		\end{tabular} &
\begin{tabular}[c]{@{}c@{}} \textbf{2.77}		\end{tabular} &
\begin{tabular}[c]{@{}c@{}} {\textbf{0.0034}} \end{tabular}
\\ \hline \hline

\begin{tabular}[c]{@{}c@{}} \\[-1em] Cartographer \end{tabular} & 
\begin{tabular}[c]{@{}c@{}} B		\end{tabular} & 
\begin{tabular}[c]{@{}c@{}} 142.97		\end{tabular} & 
\begin{tabular}[c]{@{}c@{}} 16.57		\end{tabular} & 
\begin{tabular}[c]{@{}c@{}}  0.0093 	\end{tabular}  &
\begin{tabular}[c]{@{}c@{}} D		\end{tabular} & 
\begin{tabular}[c]{@{}c@{}} 57.74		\end{tabular} & 
\begin{tabular}[c]{@{}c@{}} 4.65		\end{tabular} & 
\begin{tabular}[c]{@{}c@{}}  0.0137 	\end{tabular} 
\\ \hline
\begin{tabular}[c]{@{}c@{}} \\[-1em] RPV-SLAM \end{tabular} &
\begin{tabular}[c]{@{}c@{}} B	\end{tabular} & 
\begin{tabular}[c]{@{}c@{}} 12.91		\end{tabular} & 
\begin{tabular}[c]{@{}c@{}} 1.51		\end{tabular} & 
\begin{tabular}[c]{@{}c@{}} {\textbf{0.0009}} \end{tabular} & 
\begin{tabular}[c]{@{}c@{}} D	\end{tabular} & 
\begin{tabular}[c]{@{}c@{}} 5.67		\end{tabular} & 
\begin{tabular}[c]{@{}c@{}} 1.49		\end{tabular} & 
\begin{tabular}[c]{@{}c@{}} {0.0016}			\end{tabular} 

\\ \hline
\begin{tabular}[c]{@{}c@{}} \\[-1em] Proposed method \end{tabular} &
\begin{tabular}[c]{@{}c@{}} B		\end{tabular} & 
\begin{tabular}[c]{@{}c@{}} {\textbf{9.58}}		\end{tabular} & 
\begin{tabular}[c]{@{}c@{}} {\textbf{1.23}}		\end{tabular} & 
\begin{tabular}[c]{@{}c@{}} {0.0011}			\end{tabular} &
\begin{tabular}[c]{@{}c@{}} D		\end{tabular} & 
\begin{tabular}[c]{@{}c@{}} \textbf{4.69}		\end{tabular} & 
\begin{tabular}[c]{@{}c@{}} \textbf{0.9}		\end{tabular} & 
\begin{tabular}[c]{@{}c@{}} \textbf{0.001}			\end{tabular}  
\\ \hline

\end{tabular}

\end{center}
\end{table*}

\subsection{SLAM Trajectory Results} \label{trajectory_results}
First, an ablation study was conducted on Sequence B of the dataset (see Table \ref{tab:table_result_ablation}). Then, the proposed method was compared with the competing methods, i.e., Google Cartographer \cite{hess_real-time_2016} and RPV-SLAM \cite{kang2021rpv} (see Table \ref{tab:table_result_all}). For all of the comparisons, we have used absolute trajectory error (ATE) \cite{sturm_benchmark_2012}, relative trajectory error (RTE) \cite{geiger_are_2012}, and relative rotation error (RRE). For RRE and RTE, we have averaged these measurements over sub-trajectories with a length of 100, 200, ..., and 800m.

The ablation study aims to find the optimum value for $\theta$ in Algorithm \ref{alg:cdm} and also to evaluate the effectiveness of depth association and depth estimation modules. Bi-interpolation described in \cite{kang2021rpv} has been used as a rival to the depth estimation module.

According to Table \ref{tab:table_result_ablation}, in the absence of the depth association module, i.e., the first three rows, RPV-SLAM and the proposed method show similar performance in terms of both ATE and RTE, yet significantly better than Cartographer. Further, it proves that the bi-interpolation and the depth estimation module have no superiority over each other when there is no depth association module. On the other hand, considering different values of $\theta$, bi-interpolation and depth estimation module illustrate a similar behavior, where increasing $\theta$ leads to worse ATE and RTE. Moreover, the best accuracy for both densification methods was obtained using $\theta=2$.

As Table \ref{tab:table_result_ablation} suggests, when the depth association module was utilized (with $\theta$), more accurate results were obtained. In addition, the depth estimation module outperforms bi-interpolation in both ATE and RTE, given identical values for $\theta$. Regardless of selected depth densification methods, $\theta=2$ shows the best performance in both ATE and RTE. In conclusion, we can attain the best performance when the depth estimation module and $\theta=2$ are used. It means the depth estimation and depth association module have contributed to the improvement of the SLAM performance in both ATE and RTE. 

Table \ref{tab:table_result_all} shows the results of our best setting (depth estimation module with $\theta=2$) in comparison with Google Cartographer \cite{hess_real-time_2016} and RPV-SLAM \cite{kang2021rpv}. Google Cartographer is a LiDAR-centric SLAM that is also equipped with IMU. Overall, relatively poor results are obtained in residential areas due to unfavorable illumination conditions such as shadows. As expected, the parking lots sequence produced the best ATE performance because of its shorter length and lower scene complexity. 
Furthermore, the largest improvement in ATE was achieved in Sequence C due to the relatively shorter LiDAR ranges in residential areas.

To conclude, HDPV-SLAM produced the best results compared to Cartographer and RPV-SLAM over the test sequences. Although the performance of HDPV-SLAM varies depending on the sequence, the other SLAM systems follow a similar pattern in their performances as well. 

\subsection{Discussion}
As seen in Table \ref{tab:table_result_ablation},  the proposed technique outperforms the alternatives in terms of ATE. ATE is a metric that compares the entire trajectory to the ground truth and handles the form matching between them, indicating that the proposed method preserves the shape more effectively than the other methods. Furthermore, in three-quarters of the dataset, the RTE and RRE of the proposed technique are superior to those of the competing methods, and in one-quarter, they are second best by a slight margin. All of these comparisons demonstrate that the proposed strategy is superior.

\section{Conclusions}	\label{conclusion}
In conclusion, we present the novel HDPV-SLAM system that uses a depth estimation module and a depth association module. The depth estimation module generates depth information for panoramic images adapting a perspective model as its baseline. Furthermore, the depth association module leverages all the possible DL-based depth information in keyframes and non-keyframes. However, the proposed method has some limitations as well. We noticed YUTO MMS dataset has dynamic scene objects like cars, buses, people, etc. The SLAM system suffers from these dynamic objects that degrade its performance. As a future work, the visual features coming from these dynamic objects can be filtered out to prevent errors. Another limitation is that the depth association module uses a rule-based algorithm for updating map points. For future investigations, semantic information can be utilized for more precise validation of the matched visual feature and map point pairs.

\section*{Acknowledgement}
This initiative, entitled "3D Mobile Mapping Using Artificial Intelligence," is funded by Teledyne Optech and the Natural Sciences and Engineering Research Council of Canada (NSERC) Collaborative Research Development (CRD).

\printbibliography

\end{document}